\documentclass[journal]{IEEEtran}
\usepackage{cite}
\usepackage{amsmath,amssymb,amsfonts}
\usepackage{algorithmic}
\usepackage{graphicx}
\usepackage{textcomp}
\usepackage{hyperref}
\usepackage{multirow}
\usepackage[dvipsnames,table,xcdraw]{xcolor}
\usepackage{array}
\usepackage{xr}
\usepackage{cleveref}

\newcolumntype{P}[1]{>{\raggedright\arraybackslash}m{#1}}
\usepackage[nolist,nohyperlinks]{acronym}

\def\BibTeX{{\rm B\kern-.05em{\sc i\kern-.025em b}\kern-.08em
    T\kern-.1667em\lower.7ex\hbox{E}\kern-.125emX}}

\begin{document}

\title{Modality vs. Morphology: A Framework for Time Series Classification for Biological Signals}

\author{Jordan Tschida, \textit{IEEE Member}, Matthew Yohe, Edward Kane, \textit{IEEE Member}, Gavin Jager, Emma J. Reid, \textit{IEEE Member}, Tony G. Allen, Mark Story, \textit{IEEE Member}, Leanne Thompson, Joe Hoskins, Brandon Schreiber, Stan Seiferth, Scott Dolvin, David Cornett, \textit{IEEE Senior Member}

\thanks{Notice: This manuscript has been authored by UT-Battelle, LLC, under contract DE-AC05-00OR22725 with the US Department of Energy (DOE). The US government retains and the publisher, by accepting the article for publication, acknowledges that the US government retains a nonexclusive, paid-up, irrevocable, worldwide license to publish or reproduce the published form of this manuscript, or allow others to do so, for US government purposes. DOE will provide public access to these results of federally sponsored research in accordance with the DOE Public Access Plan (https://www.energy.gov/doe-public-access-plan). This research used resources from the ORNL Research Cloud Infrastructure at the
Oak Ridge National Laboratory, which is supported by the Office of Science of the U.S. Department of Energy under Contract No. DE-AC05-00OR22725. }

\thanks{Jordan Tschida, Matt Yohe, Edward Kane, Gavin Jager, Emma J. Reid, Tony G. Allen, Mark Story, Leanne Thompson, Joe Hoskins, Brandon Schreiber, Stan Seiferth, Scott Dolvin, and David Cornett are with Oak Ridge National Laboratory, Oak Ridge, Tennessee. (Corresponding author e-mail: tschidajl@ornl.gov).}

}

\maketitle

\begin{abstract}
Techniques for classifying biological signals are largely developed within specific modalities rather than the underlying signal structure (i.e., morphology), leading to disjointed methodologies and limited guidance across diverse signals. This work introduces a novel morphology--modality framework that connects waveform structure to a principled design of preprocessing, feature extraction, and modeling techniques for biological time-series classification. We analyze evidence across biological signals including electroencephalography, electromyography, electrocardiography, photoplethysmography, and ocular modalities including electrooculography, pupillometry, and eye-tracking, and organize according to their primary morphologies: spikes, bursts, oscillations, slow drift, and hierarchical rhythms. Across these categories, we examine how morphology influences the analytical pipeline from preprocessing to modeling and evaluation. Localized architectures favor spikes and bursts, whereas frequency and temporal models capture oscillations, and hybrid approaches integrate local and global dependencies in hierarchical signals. The associations are not absolute as biological signals often contain multiple morphologies where relevance is task dependent.  The proposed framework positions morphology as an inductive biases are useful for modeling.  We further identify challenges in cross-subject generalization, morphology-preserving data augmentation, multimodal fusion, and evaluation practices. By linking modality-specific constraints with waveform structure, this review provides guidance for selecting and evaluating pipelines across biological signals. We argue that morphology provides a useful cross-modal lens for developing biological time-series classification systems that are more generalizable, interpretable, and physiologically grounded.

.
\end{abstract}


\section{Introduction}
\label{sec:introduction}

\begin{figure}
    \centering\includegraphics[width=1\linewidth]{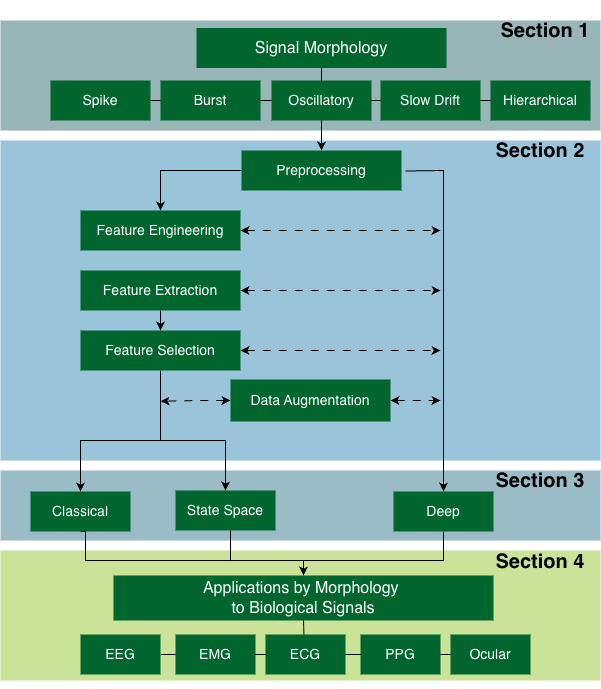}
    \caption{Dual-framework linking morphology to preprocessing, modeling, and modality specific applications for morphology-aware biological TSC.}
    \label{fig:framework}
\end{figure}

Biological signals such as electroencephalography (EEG), electrocardiography (ECG), electromyography (EMG), photoplethysmography (PPG), and ocular dynamics (electrooculography (EOG), pupillometry, eye-tracking) provide temporal patterns for diagnosis and monitoring ~\cite{wang2022systematic, antonenko2010using, lopez2023biomedical}. Time series classification (TSC) (e.g., seizure detection, sleep staging, arrhythmia detection) is challenging because biological signals are often nonstationary, highly individual, and vulnerable to noise from motion artifacts, electrode displacement, or environmental interference~\cite{swapna2022bio, sweeney2012artifact}. 

These modalities differ in their temporal structure and spectral composition. Morphologies can be broadly categorized into five recurring structures: spikes, bursts, oscillatory cycles, slow drifts, and hierarchical patterns. Spikes are isolated, high-amplitude, short-duration deflections, such as EEG epileptic discharges or ocular sharp saccadic spikes. Bursts are clusters of spiky activity occurring in rapid succession reflecting transient events like EMG muscle activations or rapid eye blinks. Oscillations capture repeating rhythms, examples include ECG cardiac cycles, PPG pulse cycles, or EEG alpha and beta bands. Slow drifts describe gradual baseline changes over time, which are commonly observed in ocular signals or as ECG and PPG artifacts. Hierarchical are  multiscale structures where local waveforms are nested within larger cycles, including the ECG P–QRS–T complex within each cardiac cycle or EEG oscillatory bands. EEG exhibits multiple morphologies including transient spikes and bursts (epileptic discharges), sustained oscillations (alpha, beta, and theta bands), and hierarchical cross-frequency coupling. 

Traditional classification approaches rely on handcrafted features and classical machine learning, offering interpretability and robustness with limited data and computational resources. These methodologies rely on domain expertise and struggle to generalize beyond the conditions or populations for which they were designed. Deep learning methods learn representations directly from raw signals: RandOm Convolutional KErnel Transform (ROCKET) captures local patterns, state-space models model temporal dependencies, recurrent and attention-based methodologies capture long-range dependencies. 

\begin{figure}
    \centering
    \includegraphics[width=1\linewidth]{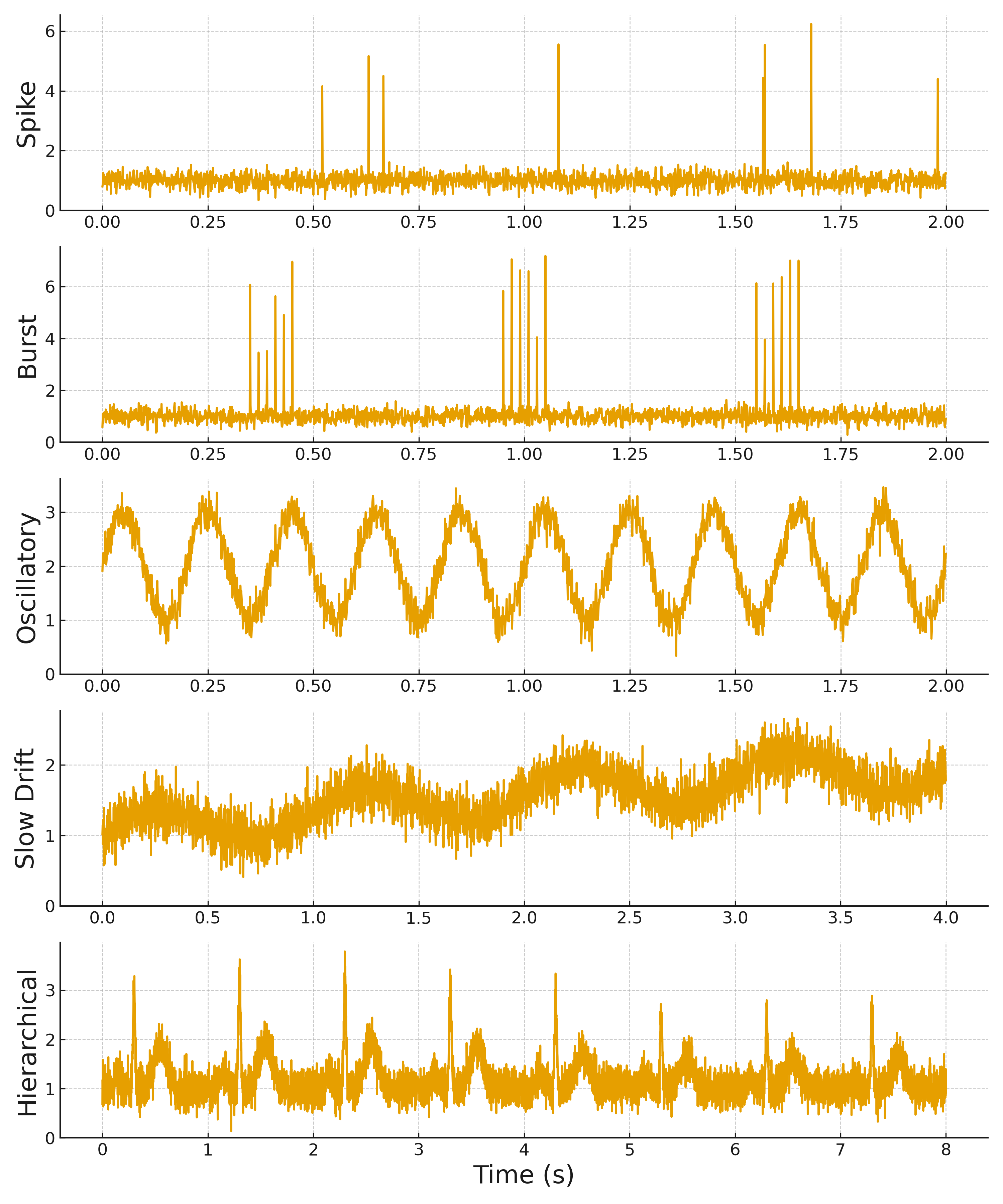}
    \caption{Generalized morphologies considered in this review: spikes, bursts, oscillations, slow drift, and hierarchical patterns. Waveforms illustrate morphological behavior rather than specific biological signals.}
    \label{fig:morphology}
\end{figure}

In this review, we introduce a dual framework that integrates (Figure \ref{fig:framework}) modality and morphology oriented perspectives. The modalities examined here include EEG, EMG, ECG, PPG, and ocular signals.  The modality perspective emphasizes considers their distinct characteristics, preprocessing requirements, and modeling challenges. The morphology perspective organizes signals by waveform structure:  spike, burst, oscillation, slow drift, hierarchical (Figure \ref{fig:morphology}). Integrating modality and morphology enables cross domain comparison and highlights techniques that may transfer across signals with shared structure.

\section{Data Techniques}
\label{sec:data}
Preprocessing transforms raw data into usable presentations while preserving the discriminative morphological structure for downstream modeling. Biological signal datasets are often small, expensive to annotate, and highly subject dependent with complications from noise, irregular sampling or missing data, and inconsistent sequence length. Incorrect data processing can distort or remove clinically meaningful structure, introduce artificial patterns, or amplify inter-subject differences causing reduced model performance. Signal distortions affect morphologies differently: artifacts mask spikes and bursts, baseline drift obscure oscillations, calibration drift and sensor artifacts confound slow drifts, and aggressive filtering regrades hierarchical rhythms. Data preprocessing provides the first inductive bias by determining what signal structure is encoded. Table \ref{tab:preprocessing} summarizes the recommendations in this Section.

\begin{table*}[]
\centering
\caption{This table operationalizes the dual morphology–modality framework by linking preprocessing and feature design to morphology (Section \ref{sec:data}).}

\begin{tabular}{|p{2.2cm}|p{2.2cm}|p{2.2cm}|p{2.2cm}|p{2.2cm}|p{2.2cm}|p{2.2cm}|}
\hline
\label{tab:preprocessing}
\textbf{Morphology} &  \textbf{Typical Modality} & \textbf{Common Artifact} &
  \textbf{Preprocessing Focus} & \textbf{Feature Engineering Priorities} &  \textbf{Feature Extraction / Selection} & \textbf{Augmentation Guidance} \\ \hline

\textbf{Spike / Burst} &
  EEG; EMG; Ocular &
  Ocular or electrical artifacts; motion noise; powerline interference &
  Remove transient artifacts (denoising); per-window or percentile normalization; high pass filtering; &
  Emphasize time-domain amplitude, burst rate, event density &
  Local kernels (ROCKET, MiniROCKET); wrapper-based selection to retain feature interactions &
  Transformation-based augmentation (jittering, cropping, and noise injection)  \\ \hline

\textbf{Oscillatory} &
  EEG; ECG; PPG &
  Baseline drift; illumination; electrode noise; motion artifacts &
  Stabilize rhythmic amplitude; cycle-based z-score; reject baseline drift &
  Frequency-domain descriptors (band power, PSD) &
  Symbolic or dictionary encodings capturing periodicity; filter-based feature selection &
  Pattern-based augmentation (modify or combine existing samples) \\ \hline

\textbf{Slow Drift} &
  Ocular &
  Baseline shift, calibration drift &
  Emphasize time-domain amplitude; Preserve baseline trends; relative-change or low-frequency normalization &
  Long-window trend and variability metrics &
  PCA/ICA for noise removal; filter selection emphasizing global variance; Local kernels (ROCKET, MiniROCKET) &
  Decomposition-based augmentation (trend and seasonality recombination) \\ \hline

\textbf{Hierarchical / Multiscale} &
  ECG; EEG; Ocular &
  Baseline wander, multiscale interference; denoising &
  Preserve inter-component ratios; multi-stage normalization &
  Combine time- and frequency-domain descriptors (wavelets, harmonics) &
  PCA/ICA; Autoencoders or hybrid selection (filter + wrapper) to balance scales &
  Generative augmentation (produce synthetic samples) \\ \hline
\end{tabular}
\end{table*}

\subsection{Preprocessing}
\label{sec:processing}
Preprocessing encodes task-specific assumptions and constrains the hypothesis space. For example, EEG bandpass filtering of 0.5--40 Hz removes ultra slow drift critical for sleep staging and high-frequency ripples required for seizure detection. Modeling performance is bounded by preprocessing choices as models cannot recover information that has been removed or distorted.  

Common preprocessing operations include interpolation of missing data (e.g., time-based, spline, and linear), noise filtering, normalization (e.g., minimum-maximum scaling or z-scoring per individual)~\cite{tawakuli2025survey}, segmentation into fixed windows, and temporal alignment. Normalization remains contested because scaling can distort signal morphology~\cite{LIMA2023normtechniques}. Batch normalization and global scaling may flatten spikes and bursts. Unit normalization or per-window scaling preserves local dynamics, but may suppress baseline drift or gradual dilation. Relative or percentile-based scaling is preferred for slow-drift morphologies like ocular signals under cognitive load because it preserves low-frequency change while controlling intersubject amplitude variation. Z-scoring or per-cycle normalization stabilizes oscillations without removing periodicity. Normalization should be selected according to the morphology and analytic goal. 

Preprocessing requirements vary by modality. EEG often requires filtering and artifact rejection to preserve spikes and oscillatory rhythms~\cite{peng2019eeg}. EMG commonly undergoes high pass filtering and denoising to preserve bursts~\cite{merletti2020tutorial}. ECG is typically preprocessed with baseline drift correction and denoising to preserve hierarchical waveforms~\cite{rai2013ecg}. PPG is vulnerable to baseline drift and motion artifacts, requiring preprocessing that stabilizes the oscillatory waveform~\cite{naraharisetti2011comparison}. Ocular signals often undergo blink interpolation and detrending to maintain ocular dynamics~\cite{lopez2023biomedical}. Importantly, artifact versus signal depends on the task: eye blinks may contaminate EEG yet convey behavioral information in ocular signals. 

\subsection{Feature Engineering}
\label{sec:feat_eng}

Feature engineering encodes discriminative information in waveform shape, timing, and rhythm. Time domain features emphasize spikes and bursts; frequency domain features capture oscillatory structure; low-frequency or trend-based features capture slow drifts; and combined representations from both time and frequency domains characterize hierarchical morphologies. 
Engineered features can be categorized into the following five domains~\cite{singh2023trends}: 

\begin{enumerate}
    \item Time domain features capture how signals evolve over time by segmenting waveforms into smaller windows and calculating basic statistics (mean, variance, skewness), higher-order descriptors (fractal dimension, detrended fluctuation), and autoregressive coefficients. These features provide interpretable summaries of spikes or bursts (e.g., EEG, EMG, ocular) and gradual baseline trends in slow drift signals (e.g., ocular), but may overlook frequency-specific oscillations. 
    
    \item Frequency domain features characterize oscillatory structure by decomposing signals into sinusoidal components. Techniques such as fast Fourier transform, power spectral density, and band power measures (e.g., alpha, beta, and delta rhythms in EEG) reveal periodicity and resonance phenomena. These features suit  oscillatory signals (e.g., EEG, ECG, PPG), but provide less temporal information for nonstationary events. 
    
    \item Decomposition techniques separate signals into constituent components using methodologies such as wavelet transforms, empirical mode decomposition, and adaptive Hermite decomposition. These approaches separate complex signals into components at different scales, supporting feature extraction from nonlinear or mixed morphologies (e.g., ocular). 
    
    \item Time–frequency features combine temporal and spectral perspectives through tools such as short-time Fourier transform, continuous wavelet transform, or S-transform. These representations localize spectral content in time, making them useful for transient and nonstationary morphologies (e.g., EEG, EMG), with a tradeoff between temporal and frequency resolution. 
    
    \item Spatial features characterize relationships across sensors or electrode arrays in multichannel recordings. Approaches such as common spatial patterns and their extensions project data into spatial filters that maximize variance differences between classes. These features support multichannel EEG and EMG analysis but may require subject-specific calibration, limiting generalization.
\end{enumerate}

Automated libraries such as TSFRESH ~\cite{CHRIST201872} and TSFEL ~\cite{barandas2020tsfel} extract statistical, temporal, and frequency domain features into a fixed-length representations and support modality- or morphology-specific customization. TSFEL features embedded in self-similarity matrices can support segmentation temporal interpretation~\cite{rodrigues2022feature}. 

\subsection{Feature Extraction}
\label{sec:feat_extract}
Feature extraction reduces high-dimensional biological signals while preserving task-relevant morphology. Unlike feature engineering (Section \ref{sec:feat_eng}), 
feature extractions uses algorithmic transformations to reduce dimensionality and redundancy while balancing compression against information loss. 

Dimensionality reduction methodologies such as PCA and ICA project data into lower-dimensional representations based on variance or statistical independence~\cite{chawla2011pca}. Autoencoders extend this approach through nonlinear embeddings that compress signals across temporal scales~\cite{vincent2008extracting}.

Dictionary-based and bag-of-patterns methodologies emphasize recurring subsequences, reducing  local variability while preserving oscillatory rhythms~\cite{craven2016adaptive, jiang2020developing}. Reduced temporal precision may limit their effectiveness for spikes, bursts, and hierarchical patterns when event timing is discriminative.

Fixed or randomly generated convolutional kernels to produce discriminative features without backpropagation. ROCKET, MiniROCKET, and MultiROCKET have shown strong performance for spikes, bursts, and oscillatory morphologies~\cite{rocket, minirocket, multirocket, solana2024classification, mizrahi2024comparative}. Their speed and scalability make them valuable when data are limited or rapid deployment is required.

\subsection{Feature Selection}
\label{sec:feat_select}
Feature selection reduces redundancy and overfitting in high-dimensional data while preserving morphology-relevant information. Approaches are categorized as filters, wrappers, and hybrid methodologies~\cite{Sánchez-Maroño_Alonso-Betanzos_Tombilla-Sanromán_2007}. 

Filter methods evaluate features independently of the downstream model using statistical measures such as the Mann-Whitney U test, Pearson correlation, and ANOVA~\cite{MannWhitneyU, vsverko2022complex, ANOVA} and are well suited to oscillatory and slow-drift morphologies. Algorithms such as Relief and ReliefF provide a computationally efficient ranked list of relevant features~\cite{kira1992_relief, Kononenko1997_relieff}. Although few filters operate directly on raw time series, they are widely used after feature engineering or extraction (described in Sections \ref{sec:feat_eng} and \ref{sec:feat_extract}). Visualization tools such as RadViz support interpretability by projecting high-dimensional feature spaces into two dimensions~\cite{radviz1997}.  

Wrapper methodologies capture feature interactions that filter methodologies overlook by optimizing a subset based on model performance using methods such as genetic algorithms, sequential forward/backward selection, and particle swarm optimization~\cite{Li_Yuan_Ma_Cui_Cao_2017, El_Aboudi_wrapper_2016}, although they are computationally expensive. This is useful for spikes and bursts where individual features may have limited discriminative power while their interactions may carry diagnostic information. 

Hybrid methodologies combine filter efficiency and wrapper precision for hierarchical morphologies that spans temporal and frequency domains. Maruotto et al.~\cite{Maruotto_Ciliberti_Gargiulo_Recenti_2025} couple statistical filtering with backward sequential feature selection using both support vector machines (SVMs) and Random Forest classifiers, whereas Kathirgamanathan et al.~\cite{Kathirgamanathan_Cunningham_2020} adapt mutual information for multivariate time series.

\subsection{Data Augmentation}
\label{sec:augemtation}
Data augmentation improves the generalization of time-series classification models when there is limited labeled data or high cross subject variability by reducing overfitting and increasing robustness to signal distortions and noise. Augmentation strategies align naturally with morphology: transformations preserve spikes and bursts, pattern-based methodologies reinforce oscillatory rhythms, decomposition manipulate slow drifts, and generative approaches model complex hierarchical structures. 
Gao et al.~\cite{Gao2025} categorize augmentation strategies into four classes: 
\begin{enumerate}
    \item Transformation-based methodologies apply jittering, scaling, or alterations to the time or frequency characteristics; these are broadly effective but may introduce unrealistic distortions when applied too aggressively~\cite{Um2017DataAug, leguennec2016dataaug, ZhangTFC}. 

    \item Pattern-based methodologies generate new samples by combining or modifying existing patterns, potentially improving performance but also risking amplifying biases in the data~\cite{Iwana2021DataAug}.

    \item Decomposition-based methodologies separate signal components such as trend and seasonality, augment them separately, and recombine them, providing interpretable control over the manipulated features but often having limited performance~\cite{Iwana2021DataAug, Gao2025}. 

    \item Generative methodologies leverage models like generative adversarial networks or variational autoencoders to produce synthetic time-series samples~\cite{Morizet2022DataAug}. 
\end{enumerate}

Counterfactual generation creates minimally perturbed new samples that shift model predictions, helping define decision boundaries and improve robustness and generalizability~\cite{holligTSEvo2022}; however, these are computationally expensive and may produce unrealistic artifacts if applied too aggressively.

\section{Modeling Techniques}
\label{sec:modeling}

Model choice depends on dataset size, annotation, computational resources, and the signal's morphology: local feature-based models favor spikes or bursts, temporal models capture oscillations and slow drifts, and hybrid architectures integrate both local and global context in hierarchical signals. These associations represent inductive biases rather than rigid requirements. Table \ref{tab:roadmap} synthesizes this framework.

\begin{table*}[]
\centering
\caption{Mapping of waveform morphology to modeling approach and deployment constraints integrating Sections \ref{sec:diss_eval} and \ref{sec:diss_interp}}
\label{tab:roadmap}
\begin{tabular}{|p{2.2cm}|p{2.2cm}|p{2.2cm}|p{2.2cm}|p{2.2cm}|p{2.2cm}|p{2.2cm}|}
\hline
\textbf{Morphology} & \textbf{Typical Signals} & \textbf{Lightweight (CPU)} & \textbf{Transformer (GPU)} &  \textbf{Evaluation \newline  Metrics} & \textbf{Model Selection Guidance}  & \textbf{Interpretability}\\ \hline

\textbf{Spikes/Bursts} & EEG, EMG, EOG &  ROCKET/ MiniROCKET; SVM; CNN & Time-Series Transformer; CNN--Transformer hybrid & Recall; precision; detection latency; false positive rates & Lightweight for strict latency; Transformer if events are context or long sequences are required  & Wavelet or kernel attribution; Counterfactual Spike Insertion/Removal\\ \hline

\textbf{Oscillatory Rhythms} & EEG, ECG, PPG & CNN; State-Space & Transformer; State-Space--Transformer hybrid & Frequency stability; phase error; beat-to-beat accuracy; HRV error; calibration drift & Lightweight for wearables; Transformer for nonstationary  or fused rhythms & Cycle-synchronous attention mechanisms; SHAP/LIME\\ \hline

\textbf{Slow Drift} & Pupillometry & RNN or filter-based; CNN & Transformer; Transformer hybrids &  Drift error (RMSE); baseline recovery error; session-to-session stability & Lightweight for smooth, short windows; Transformer when long-range dependencies or mixed drift + events & Residual analysis, detrending diagnostics, and baseline shifts perturbation\\ \hline

\textbf{Hierarchical} & ECG, multimodal &  Multiscale CNN & Hierarchical Transformer; State-Space--Transformer hybrid & Macro-F1; cross-scale consistency; alignment error across sensors & Lightweight for shallow hierarchies; Transformer when global context or cross-scale coupling & CNN saliency for local waveforms, attention or state-space reconstructions for global rhythms\\ \hline
\end{tabular}
\end{table*}

\subsection{Evaluation Metrics}
\label{sec:diss_eval}
Evaluation metrics remain inconsistent~\cite{wang2022systematic}. Accuracy and F1 scores dominate evaluation but overlook temporal and morphological performance. Time-aware metrics capture responsiveness over time and morphology-specific metrics capture waveform fidelity, enabling more consistent benchmarking.

\subsection{Interpretability}
\label{sec:diss_interp}
Clinical deployment requires safe and interpretable models that provide evidence that predictions rely on physiologically meaningful features rather than artifacts. Localized interpretability tools such as wavelet-based feature attribution, ROCKET explanations, or counterfactual spike insertion/removal reveal whether the model depends on physiologically meaningful information for spike and burst signals. Cycle-synchronous interpretability using attention mechanisms or phase-aligned saliency maps evaluates whether P, QRS, and T waves are consistently weighted while methods such as SHAP and LIME further evaluate whether models are sensitive to rhythms rather than artifacts for oscillatory signals. Residual analysis, detrending diagnostics, and perturbations that simulate baseline shifts help distinguish physiological changes from calibration error for slow drifts. For hierarchical patterns, hybrid interpretability combines CNN saliency for local waveforms with attention or state-space methods to contextualize their contribution to global rhythms. In multimodal fusion, attribution methods can reveal how complementary modalities and morphologies contribute to a decision. Aligning interpretability techniques with morphology enables physiologically grounded model reasoning.

\subsection{Classical Machine Learning Methodologies}
\label{sec:classical_ml}
Classical machine learning methodologies remain competitive when datasets are small, labels are scarce, or interpretability is essential, but rely on handcrafted features ~\cite{balli2010classification}. 

Model performance is shaped by the morphology and noise profile because each methodology imposes distinct inductive biases. Spike and burst activity in EEG and EMG are effectively modeled using SVMs, which handle high-dimensional data and sparse features~\cite{sha2020knn, lofhede2008comparing, pfammatter2019automated}. Oscillatory cycles observed in ECG and PPG align with logistic regression when features are engineered to capture smooth, correlated patterns~\cite{aspuru2019segmentation, haddad2021continuous}, as well as distance-based classifiers like k-nearest neighbor, which exploits the similarity of the waveform~\cite{toulni2021ecg, saini2013qrs}. For signals with slow drift or irregular noise, ensemble methods like random forests provide robustness by averaging across heterogeneous feature sets, perform implicit feature selection, and capture nonlinear dependencies~\cite{wang2022random}.

\subsection{State-Space and Hybrid Sequence Models}
Hidden Markov Models (HMMs) represent biological signals through latent physiologically meaningful states~\cite{mehari2023towards, zou2025rhythm}, capturing oscillatory cycles in PPG or ECG~\cite{mehari2023towards, zou2025rhythm}, mid-frequency oscillations in EEG~\cite{manomaisaowapak2021granger}, and fast spike transitions in EMG~\cite{xi2021simultaneous}. Their latent-state representation also corresponds to hierarchical morphologies, such as the ECG P-QRS-T complex. This layered representation enables SSMs to accommodate slow drifts while remaining responsive to discriminative bursts.

\subsection{Deep and Attention-Based Architectures}
\label{sec:deep_models}
Deep and attention-based architectures scale high-dimensional data while capturing multiscale temporal patterns. CNNs capture local morphological patterns when discriminative information is localized including spikes, bursts, oscillatory signals. Recurrent architectures such as LSTMs capture long-range temporal dependencies, making them well-suited for modeling slow drifts, state transitions, and other non-stationary dynamics. This capability is critical for modalities like ECG where arrhythmia detection depends on rhythm history; PPG which requires tracking baseline wander; and ocular where gradual eye position shifts must be represented over time. Transformer and related attention-based architectures are effective at modeling hierarchical morphologies like multiband EEG rhythms~\cite{narula2021detection}, oscillatory rhythms in ECG and PPG~\cite{ezzat2024ecg}, and ocular movements~\cite{qin2025emmixformer}. Hybrid architectures combine these inductive biases to model local and global temporal structure.

Deep models face challenges because their effectiveness often depends on large, labeled datasets and interpretability remains limited, complicating clinical adoption and generalization. Their computational requirements can limit deployment in low-resource environments or real-time settings. 

\section{Morphology Based Applications}
\label{sec:bio_signal_morphology}
Applications span clinical and nonclinical domains, with morphology informing task utility. EEG and EMG spikes and bursts support seizure detection, motor-intention decoding, and muscle monitoring, whereas ECG hierarchical and PPG oscillatory rhythms  enable arrythmia detection and sleep apnea screening~\cite{cai2023EEGDict, li2022classification, JiangCNNApnea, tran2024eegssmleveragingstatespacemodel}. Pupillometry slow drifts reflect attention patterns, cognitive workload, and emotional state that provide insights for psychology~\cite{DasPupillometry, Liu2022, rutkowska2024optimal}. This section organizes modality applications by morphology with Appendix Table \ref{tab:integrated_framework} summarizing each application and modeling choice. 

\subsection{Biological Signals}
\label{sec:bio_signals}
\subsubsection*{EEG}
EEG measures brain activity at millisecond resolution through scalp electrodes, enabling applications such as sleep staging, seizure detection, cognitive workload, motor imagery, and emotion classification~\cite{roy2019deep, singh2023trends}. EEG provides high temporal resolution but lacks spatial resolution and is highly susceptible to physiological artifacts (e.g., cardiac activity, eye movements) and environmental interference (e.g., ambient sounds, power-line interference, electrical noise)~\cite{roy2019deep, singh2023trends}. Morphologically, EEG combines transient spikes and bursts (epileptic discharges) with sustained oscillatory rhythms across multiple frequency bands (alpha, beta, delta, gamma, theta) and hierarchical organization through nested cross-frequency coupling~\cite{kumar2012analysis}. Because the dominant morphology shifts with cognitive or clinical context, EEG is a highly nonstationary and participant-dependent signal, requiring models that adapt to evolving temporal structures.

\subsubsection*{EMG}
EMG measures electrical activity in muscles at millisecond resolution through surface or needle electrodes, enabling applications such as gesture recognition, fatigue detection, and decoding motor intentions for prosthetic control or gesture-based control systems~\cite{mills2005basics, farago2022review}. EMG waveforms are dominated by short bursts of noisy spikes corresponding to muscle contractions and are highly nonstationary, vary across participants, and are often corrupted by muscle cross-talk, motion, or electrode noise~\cite{farago2022review}. 

\subsubsection*{ECG}
ECG measures cardiac electrical activity at millisecond resolution through electrodes placed on the chest, supporting applications such as arrhythmia detection, stress monitoring, and myocardial infarction diagnosis~\cite{mirvis2001electrocardiography}. ECG is inexpensive, noninvasive, and physiologically interpretable but can be corrupted by motion artifacts, electrode placement, and inter-person variability ~\cite{mirvis2001electrocardiography}. The ECG waveform exhibits both oscillatory and hierarchical structure, rhythmic repetition of cardiac cycles composed of P waves, QRS complexes, and T waves~\cite{HurstWavesNaming, ECGwaves}. 

\subsubsection*{PPG}
PPG measures blood-volume variation at millisecond resolution via optical sensors placed on the skin and is widely employed for heart rate and blood pressure estimation, affective computing for stress and emotion recognition, and detection of atrial fibrillation and sleep apnea~\cite{castaneda2018review}. PPG is inexpensive, portable, and well-suited for continuous health monitoring; however, it is highly sensitive to motion artifacts and sensor variability~\cite{castaneda2018review}. Morphologically, PPG waveforms are characterized by smooth oscillatory rhythms with a systolic peak followed by a diastolic phase, with slow baseline drift as a secondary component.

\subsubsection*{Ocular}
Ocular signals capture multiple aspects of eye activity at millisecond resolution and support applications such as wheelchair control, sleep-stage classification, and assessments of cognitive load, deception, and decision-making ~\cite{lopez2023biomedical, sirois2014pupillometry, carter2020best}. These signals are acquired through both electrophysiology and optical imaging: EOG measures cornea-retinal potential differences through electrodes placed around the eyes ~\cite{lopez2023biomedical}, pupillometry quantifies pupil-diameter changes using infrared photography or video-based eye tracking systems ~\cite{sirois2014pupillometry}, and eye-tracking systems map gaze position to reveal attention and decision processes ~\cite{carter2020best}. Morphologically, EOG waveforms exhibit burst-like deflections for saccades and sharp drops for blinks; pupillometry produces slow fluctuations with phasic dilations reflecting autonomic or cognitive responses; and eye-tracking alternates between flat fixation plateaus and spiked transitions marking saccades. This combination of modalities (i.e., electrical and optical) and morphologies (e.g., bursts, slow drifts, and hierarchical patterns) underscores the value of multimodal integration; however, they are sensitive to lighting, calibration drift, and motion artifacts~\cite{lopez2023biomedical, sirois2014pupillometry, carter2020best}.

\subsection{Spikes and Bursts}
\label{sec:spike_burst}
Spikes and bursts are brief, high-amplitude events that reflect discrete neural or muscular activations. Their temporal sparsity and high energy concentration require architectures capable of detecting short-duration events with millisecond precision while maintaining robustness to noise and variability. 

\subsubsection*{EEG}
Traditional machine learning models offer competitive results for EEG classification; decision trees, k-nearest neighbor, and Bayesian classifiers achieved 81\% accuracy on a dataset of 121 participants for seizure detection~\cite{KaratasClassicalEEG2025}; demonstrating that engineered features can effectively capture seizure-related spike and burst morphology.

Dictionary learning methodologies improved seizure detection baselines by 8\% on the CHB-MIT and Bonn datasets by mitigating challenges associated with inter-person and inter-device variability~\cite{cai2023EEGDict}. This approach demonstrates the ability of dictionary learning to capture seizure-related patterns but may struggle to scale. 

Hybrid CNN-Transformer models such as EEG Conformer and EEGformer combine convolutional layers to extract local temporal–spatial patterns with self-attention to capture global temporal features, achieving state-of-the-art performance across EEG decoding and brain activity benchmark datasets~\cite{eegConformer, EEGFormer2023}. This hybrid design demonstrates the value of combining local inductive biases with global context.

\subsubsection*{EMG}
\label{sec:emg}
Classical machine learning approaches rely on engineered time-frequency features such as spectral entropy and mean frequency. SVMs have been used for both classification and regression tasks. For gait disorder classification from EMG signals collected from 37 individuals, an SVM achieved 68\% accuracy using manually extracted features, whereas a CNN achieved 92\% accuracy on detecting a healthy gait from impaired gait~\cite{fricke2021evaluation}. For SVM basedankle torque prediction, high-density EMG improved accuracy compared with traditional isometric recordings; however, accuracy improved most when EMG features were augmented with joint kinematics (angle and velocity), reducing errors in complex movements such as gait stance by over 60\%~\cite{kizyte2023influence}. 

State space models explicitly capture the temporal evolution of EMG bursts. A Bayesian Hierarchical Dirichlet Process HMM achieved 96\% accuracy across 17 hand and wrist movements, outperforming a baseline HMM~\cite{wen2021human}; however, such methodologies may have challenges scaling to the high dimensionality of multichannel EMG and complex motor tasks. 

Deep learning models capture EMG burst morphology directly with CNNs and CNN-LSTMs. CNNs apply convolutional filters across electrode channels or spectrograms, capturing localized spatial and frequency patterns that correspond to muscle activity. CNN-LSTMs extend this capability: the CNN extracts local spatial or short-term features, whereas the LSTM models sequential dependencies across time, making them particularly advantageous to gestures involving muscle patterns. A CNN applied to high-density EMG predicted cumulative spike trains without condition-specific motor unit templates, achieving correlation $>0.96$, root mean square error (RMSE) $<7\%$) while generalizing across datasets~\cite{WenEMG-CNN}. A CNN-LSTM applied to spectrograms of 300 participants performing five hand gestures achieved accuracy of 90\%~\cite{lopez2024cnn}. A lightweight CNN achieved 91\% accuracy on NinaPro DB1 using time-frequency features and concatenated CNN blocks, improving the baseline by 25\% while requiring fewer parameter~\cite{qureshi2023e2cnn}. 

Transformer-based models extend EMG modeling to long-range dependencies and distributed inter-muscle interactions. A compact Vision Transformer (ViT) applied to 65 isometric hand gestures across 19 participants achieved accuracies above 90\% ~\cite{montazerin2023transformer}. Another ViT introduced Blind Source Separation for separating mixed signals and was evaluated on both the DB1 and DB2 datasets; the best model achieved 96\% accuracy on DB1 and 92\% on DB2, outperforming baselines by 6\%~\cite{dere2023novel}. Hybrid convolutional-attention models such as EMG-TransNN-MHA have reached 96\% accuracy on the Upper-Limb Gesture dataset ~\cite{joby2024emg}, whereas the Dual Transformer Network has predicted hip, knee, and ankle joint kinematics, achieving $1.1827^\circ$, $1.4312^\circ$, and $0.8113^\circ$ average RMSE while highlighting physiologically meaningful muscle weightings~\cite{wang2025dual}. 

\subsubsection*{EOG}
HMMs applied to smooth pursuit decoding reached accuracy of 91\% in adults and 53\% in children, outperforming distance-based heuristic baselines and approaching human-level annotation reliability ~\cite{kim2020hidden}. This study illustrates how probabilistic modeling can extract clinically meaningful information from low-dimensional ocular signals. 

CNNs operating directly on raw EOG detected blinks and eye movements with 90\% accuracy across four directions~\cite{ravichandran2021electrooculography}. Hybrid SE-ResNet–Transformer models applied to EOG improved sleep staging, achieving macro-F1 of 75\% on SleepEDF and SHHS datasets with notable gains in REM detection ~\cite{maiti2024enhancing}. These results demonstrate how convolutional models capture transient ocular events, while hybrid architectures incorporate broader temporal context for EOG-based sleep staging.

\subsection{Oscillatory}
Oscillatory signals exhibit rhythmic structure driven by underlying physiological cycles, such as cardiac, respiratory, or cortical rhythms. These waveforms are defined by phase, amplitude, and frequency dynamics. Modeling approaches must therefore preserve spectral stability while remaining sensitive to deviations in rhythm and phase. This morphology favors frequency-domain representations and models that encode band-limited temporal dynamics over time.

\subsubsection*{EEG}
EEG classification with traditional machine learning models relies on handcrafted frequency band features, offering competitive results. Wavelet-based descriptors combined with random forests achieve 91\%--97\% accuracy on sleep stage classification~\cite{da2017single}, whereas statistical, spectral, and temporal features extracted with TSFEL support schizophrenia detection with 92\% accuracy using logistic regression on a balanced dataset of 28 participants~\cite{hussain2021evaluating}. State-space models extended EEG analysis by capturing transitions between latent brain states. HMMs achieved 77\% accuracy for classifying sleep stages on the Sleep-EDF dataset ~\cite{pouliou2025new}.

CNN applied to spectrograms and power spectral density achieved 96\% accuracy on the Emotion EEG dataset (SEED) and 93\% on the Database for Emotion Analysis using Physiological Signals (DEAP) for emotion recognition~\cite{MahmoudEEGemotion2023}. Multi-Band EEG Transformer (MEET) decomposed EEG into frequency bands and applied temporal and spatial attention, achieving 96\%--99\% accuracy on the SEED, SEED-IV, and WM benchmark datasets, while producing interpretable band-specific attention maps corresponding to relevant brain regions~\cite{MEET2024}. 

\subsubsection*{ECG}
Classical machine learning approaches remain effective because ECG exhibits regular, clinically interpretable morphology. An SVM classifier trained on waveform morphology achieved 62\% accuracy, increasing to 97\% when morphological and frequency features were fused across 540 ECG samples across nine arrhythmia classes from the MIT-BIH, LTAFDB, and ProSim 2 databases ~\cite{li2022classification}. 

For cuffless blood pressure estimation, a regression model employing TsFRESH was evaluated on 69 participants in ambulance settings; the best model achieved mean absolute errors of 12.81 ± 2.66 mmHg (SBP) and 8.12 ± 1.80 mmHg (DBP); those values improved to 6.93 ± 4.70 mmHg and 7.13 ± 4.48 mmHg after calibration~\cite{kochev2020novel}. Although a regression task, this study demonstrates the utility of automated ECG feature extraction while highlighting the importance of calibration and personalization.

State-space models extend ECG analysis by capturing long-range temporal dynamics and transitions in rhythm. ECGMamba, an SSM-based framework for ECG classification, achieved accuracy above 90\% on the PTB-XL and CPSC2018 datasets with strong F1 and AUC scores relative to traditional baselines~\cite{qiang2024ecgmamba}. Similarly, WildECG, a self-supervised SSM pretrained on 275,000 ECG segments, demonstrated strong generalization, achieving a Concordance Correlation Coefficient of 0.356 on AVEC-16, 97\% accuracy on WESAD, and 95\% accuracy on stress detection in SWELL-KW~\cite{AvramidisECGSSM2024}. 

Hybrid and deep learning models combine local feature extraction with global sequence modeling. CNN-based pipelines such as Fuzz-ClustNet achieved 98\% accuracy on the MIT-BIH and 96\% accuracy on PTB Diagnostic datasets by combining CNN-based feature extraction with fuzzy clustering~\cite{kumar2023fuzz}. CNN-LSTM hybrids extend this approach by coupling spatial features with temporal recurrence, achieving 99\% accuracy after class balancing and ensemble methodology on the PTB Diagnostic and MIT-BIT Arrhythmia datasets~\cite{rai2022hybrid}; for more information on ensemble methodologies, refer to Appendix Section \ref{sec:ensemble}. 

Transformer-based architectures push performance further. A Constrained Transformer Network achieved 79\% accuracy and improved embedding separability across nine arrhythmia categories~\cite{che2021constrained}. A CNN-Transformer hybrid, ECG DETR, reframed arrhythmia detection as object detection, achieving 99\% accuracy on MIT-BIH and AF databases~\cite{hu2022transformer}. ECGTransForm integrated multiscale convolutions and channel recollaboration, obtaining macro-F1 scores of 94\% and 99\% on the MIT-BIH Arrhythmia and PTB Diagnostic ECG datasets~\cite{ElGhaishECGTransformer}. These results illustrate how attention-based ECG models combine local morphology with global temporal reasoning.

\subsubsection*{PPG}
Classical feature-based approaches  remain effective for modeling PPG because its oscillatory morphology can be summarized by pulse shape and heart-rate variability (HRV) features. PPG HRV-derived features have been validated as low-cost surrogates for ECG in stress detection achieving F1-scores above 0.80 and AUC values around 0.90 and 0.70 across sensors~\cite{benchekroun2022comparison}. Regression-based pulse wave analysis for continuous blood pressure estimation achieved clinical accuracy standards (sensitivity = 79\%; specificity = 92\%)  ~\cite{haddad2021continuous}. 

State-space methodologies extend PPG analysis by explicitly modeling temporal dynamics. PhysMamba integrates SSMs with attention block mechanisms under a Synergistic State Space Duality framework, emphasizing periodic features such as heart rate. On UBFC-rPPG, PURE, and MMPD datasets, PhysMamba achieved state-of-the-art robustness, with mean absolute errors as low as 0.45 bpm (UBFC), 0.24 bpm (PURE), and 2.84 bpm (MMPD)~\cite{yan2025physmamba}.

CNN-based pipelines exploit local pulse morphology. CNN–MLP hybrids achieved 82\% accuracy in binary stress classification on UBFC-Phys~\cite{Hasanpoor_2022}, whereas 1D-CNN-LSTM detected sleep apnea with 91\% accuracy across 59 participants~\cite{JiangCNNApnea}. Lightweight CNNs have also been optimized for  IoMT signal quality assessment by training on nearly half a million PPG segments, achieving 95\% accuracy in detecting PPG quality~\cite{sivanjaneyulu2022cnn}. 

Transformers capture long-range dependencies in PPG, supporting blood pressure and physiological-state estimation. Raw-signal Transformers estimated systolic and diastolic blood pressure and oxygen saturation on MIMIC-III,  achieving mean absolute errors of 2.4 mmHg (SBP), 1.3 mmHg (DBP), and 0.6\% SpO$_2$, all within clinical accuracy standards~\cite{chu2023non}. KD-Informer employed knowledge distillation with morphological feature fusion for efficient blood pressure waveform estimation, reducing parameters by 83\% while achieving 5.93 mmHg systolic and 3.87 mmHg diastolic errors across 467 Mindray and 20,000 MIMIC patients ~\cite{ma2023kdinformer}. Similarly, PCTN fused CNN and Transformer branches for cuffless BP estimation reaching 2–4 mmHg errors on MIMIC-III~\cite{tian2025paralleled}. Beyond vital signs, TranSenseFusers fused PPG with electrodermal activity and skin temperature via CNN–Transformer modules for stress detection, achieving 98\% accuracy and 97\% F1-score on WESAD with interpretable attention maps ~\cite{KasnesisTransSenseFusers2025}. Together, these approaches capture both local pulse morphology and longer-range oscillatory dynamics.

\subsection{Slow Drift}
Slow drift morphologies reflect gradual, low-frequency fluctuations in baseline activity that correspond to sustained physiological or cognitive states and are influenced by arousal, fatigue, or environmental conditions. Effective modeling requires architectures capable of capturing long-range temporal dependencies and distinguishing meaningful signal variation from baseline trends and artifacts.

\subsubsection*{Pupillometry}
Classical approaches extract features such as blink rate, pupil dilation variability, or fixation duration. Although these methodologies achieve reasonable performance, they may not capture complex temporal dynamics in pupil responses. 
A feature-based framework extracted over 700 temporal and frequency features, enabling an SVM to distinguish ADHD from controls with 85\% AUROC, 77\% sensitivity, and 75\% specificity~\cite{DasPupillometry}. 

Deep learning models capture nonlinear interactions between pupil dynamics and contextual influences. Bidirectional LSTMs trained on fixations, saccades, blinks, and pupil features separated luminance and accommodation effects, with residual pupil variation more strongly associated with learning outcomes than raw pupil size (AUC = 0.74)~\cite{LSTM_noncog_filter}. CNN models trained on pupil spectrograms detected PTSD with 81\% accuracy, validating frequency-domain features for psychiatric screening~\cite{taha2021detection}. %

\subsection{Hierarchical}
Hierarchical morphologies combine localized events with broader physiological states or rhythms, rapid transients embedded within slower oscillatory cycles, requiring models that capture both short- and long-range dependencies.

\subsubsection*{ECG}
DENS-ECG fuses CNN with bidirectional LSTM networks to capture both sharp deflections and slower phases, reporting F1 scores of 96\% for T-waves and 99\% for QRS waves~\cite{peimankar2021dens}. 

\subsubsection*{Eye-Tracking}
Transformer-based architectures capture long-range dependencies and multimodal eye dynamics. Event-based gaze tracking with a bidirectional relative positional attention Transformer achieved 98\% accuracy within 5 pixels and mean error of 1.14 pixels~\cite{wu2025brat}. EmMixformer combined an attention-enhanced LSTM, Transformer, and Fourier Transformer, achieving a low equal error rate on the GazeBase, JuDo1000, and EMglasses datasets~\cite{qin2025emmixformer}. 

Vision Transformers applied to pupil segmentation achieved 99\% accuracy across diverse images and surpassed CNN baselines~\cite{vayalil2024vit}. Although not a classification task, these results underscore the adaptability of Transformer architectures to ocular biosignal domains. 

Probabilistic vestibulo-ocular reflex signals fused with gyroscope data in an HHM reduced gaze-tracking error to 3.54$^{\circ}$, enabling continual calibration from natural head movements~\cite{nezvadovitz2022using}.

\subsection{Multimodal Morphological Fusion}
\label{sec:multi_modal}
Multimodal biological signal modeling integrates complementary  modalities and morphologies within a unified representation. Within the dual framework, fusion links distinct signals by integrating oscillatory rhythms, sharp spikes, and slow drifts. The challenge is preserving those complementary structures while accounting for different sampling rates, artifacts, and sensor properties and maintaining temporal synchronization.

Affect-aware models trained on courtroom videos achieved 84\% accuracy on deception detection by fusing valence-arousal features from facial, visual, vocal, and verbal cues~\cite{Mathur_2021}. Similarly, micro-movement analysis of ocular (blinks, gaze shifts) and facial cues revealed that ocular cues provided stronger indicators of deception than cues provided by facial features, with random forests achieving 85\% accuracy ~\cite{KhanDeception}. 

A multi-representation fusion of time–frequency time–space LSTM (TFTS-LSTM) improved ECG atrial fibrillation classification from 58\% to 94\% accuracy and achieved perfect accuracy in gait-based Parkinson’s detection using a single foot sensor~\cite{pham2021tfts}. 

EEGdenoiseNet, a Transformer framework for denoising EEG, combined multiscale module slice-pattern attention achieved superior RMSE than baselines~\cite{chen2024denosieformer}. Fusion of EEG and pupillometry for depression recognition improved accuracy from 90\% for EEG and 84\% for pupillometry alone to 93\% when combined~\cite{zhu2025transformer}. EEG and eye-tracking fusion with a multimodal Transformer achieved 92\% accuracy for adolescent depression detection, compared with 88\% for EEG and 76\% for eye tracking alone~\cite{zhu2025mtnet}.

EEG–EMG fusion via FAConformer combines band-specific attention and multiscale convolution outperforming baselines~\cite{xiao2024faconformer}. The Performer model introduced Shifted Patch-Based Attention for PPG-to-ECG reconstruction, enabling cardiovascular disease detection with 96\% accuracy on MIMIC-III~\cite{lan2023performer}. A hybrid CNN–Transformer for cuffless blood pressure monitoring integrated ECG, PPG, and pulse wave inputs, achieved mean errors within 1 mmHg and reduced systolic error by nearly 10\%~\cite{liu2024hgctnet}.

\section{Discussion} 
\label{sec:discussion}

This framework suggests that biological signal TSC is better characterized by the interaction between signal modality and morphology. Across EEG, EMG, ECG, PPG, and ocular signals, model performance consistently reflects alignment between inductive biases and the temporal structure of discriminative information. Localized architectures favor spikes and bursts, frequency-aware models capture oscillatory structure, recurrent and state-space approaches represent gradual temporal evolution, and hybrid architectures integrate local and global dependencies in hierarchical signals. Together, these patterns support the dual modality--morphology framework as an organizational principle for biological TSC. 

These relationships should not be interpreted as fixed mappings, some signals do not fit neatly into a single morphology. For instance, EEG exhibits spikes nested within oscillations and broader hierarchical rhythms, making it difficult to assign a dominant morphology. Ocular signals pose a similar ambiguity: they combine slow drifts related to cognitive load with spikes from blinks and illumination changes. 
Morphology is therefore better understood as a task-dependent inductive bias than as a static signal category.

Importantly, morphology--model alignment represents an inductive-bias perspective rather than a rigid prescription. The same modality may require different architectures depending on whether discriminative information is localized, periodic, slowly evolving, or distributed across temporal scales. Multimodal fusion extends this principle by combining complementary morphologies, but introduces additional challenges in synchronization, artifacts, and sensor heterogeneity.

This interpretation extends beyond model selection to the entire TSC pipeline. Jittering or scaling can distort spike and burst intervals; time warping can alter the frequency band in oscillatory morphologies; sliding-window segmentation of hierarchical data may cut across the nested structures; and detrending slow drift may remove clinically relevant baseline changes. 

Variability also occurs across (interperson) and within (intraperson) individuals due to aging, health status, and emotional state, thus altering signal properties over time \cite{wei2021variation}. Voluntary signals (e.g., gait, breathing, voice in Appendix Section~\ref{sec:misc}) add intent-driven variability; invariance may remove the discriminative information. These issues are amplified in unconstrained real-world environments, where sensor heterogeneity and motion artifacts further degrade stability~\cite{Seghier2018}. Domain-invariant representation learning has shown promise for addressing interperson variability~\cite{gu2021cross, rayatdoost2021subject}; however, such methodologies remain underutilized and computationally demanding. 

Generative methodologies may produce signals that reproduce global characteristics while distorting the morphological structure that defines discriminative features, such as amplitude deviations and rare events. Reconstruction- or likelihood-based objectives may suppress these events by favoring smoother representations. Without morphology constraints, both augmentation and generation risk amplifying biases rather than improving generalization. Morphology-informed invariance and counterfactual generation can improve robustness without suppressing meaningful physiological variation. 

The available evidence is also constrained by existing benchmarks. Large repositories such as MIMIC, UK Biobank, and MESA include multimodal data (ECG, PPG, EEG, EOG), whereas PhysioNet \cite{goldberger_physiobank_2000} curates smaller, task-specific datasets. However, most large datasets are designed for clinical and diagnostic settings. Behavioral datasets remain scarce and difficult to synchronize. 
Differences in sensors, participant populations, preprocessing, and evaluation protocols further complicate cross-study comparison. Consequently, reported performance should be interpreted within individual dataset and task contexts rather than as direct evidence of architectural superiority.

Future benchmarks should therefore span modalities, morphologies, and application contexts while incorporating standardized preprocessing, unified data structures, and controlled nuisance conditions such as motion, drift, and calibration errors. Appendix Section \ref{sec:datasets} and Appendix Table \ref{tab:dataset_table} summarize existing resources, but new community benchmarks are needed to enable rigorous cross-study evaluation and broader generalization.



Future research should also move toward multilevel representations that allow morphologies to coexist and evolve across tasks rather than assigning each signal a single dominant structure. Multimodal models provide a natural extension by integrating complementary structures across sensors, while self-supervised learning may enable transferable representations from large unlabeled datasets. Both directions require objectives that preserve physiologically meaningful temporal structure rather than suppressing it during representation learning.

Progress will additionally require morphology-aware evaluation that extends beyond aggregate classification accuracy to temporal responsiveness, robustness, waveform fidelity, and cross-subject generalization. Standardizing these measures alongside datasets and preprocessing would enable more rigorous comparison across studies.

The literature suggests that biological TSC should move beyond selecting models according to modality. Treating morphology as an additional organizing principle connects signal structure with preprocessing, representation learning, model inductive bias, multimodal fusion, and evaluation. This perspective shifts the objective from identifying a best architecture for each biosignal toward designing models whose assumptions align with the physiological structure relevant to the task.

\section{Conclusion}
Time-series classification of biological signals has progressed from handcrafted, modality-specific approaches to deep and hybrid architectures capable of modeling complex morphologies across modalities. Real-world deployment continues to expose limitations in data scale, evaluation metrics, and model interpretability. The proposed dual framework links modality-specific constraints with morphology-driven methodological choices, providing a unified perspective for biological TSC. Progress will depend on three priorities: (1) expanding and standardizing datasets, (2) developing evaluation metrics that reflect real-world temporal and clinical scenarios, and (3) designing models that balance accuracy with interpretability. Morphology-aware and self-supervised models may support transfer across modalities, although inter- and intra-individual variability continue to hinder real-world performance. The dual modality–morphology framework underscores that the goal is not to identify a single optimal model per modality, but to bridge modality-specific characteristics with morphology-based principles to achieve generalizable, interpretable, and clinically meaningful solutions.

\section*{Acknowledgment}
This work was funded by the US Defense Counterintelligence and Security Agency's Research Sciences and Innovation group.

\section*{Declaration of AI Use}
The authors used generative AI to assist in editing the manuscript for clear, concise language.

\appendix

\subsection{Signal Modality}
Across the applications reviewed, spikes and bursts are primarily represented in EEG, EMG, and EOG; oscillatory structure is prominent in EEG, ECG, and PPG; slow drift is represented by pupillometry; and hierarchical structure appears in ECG, EEG, and eye-tracking signals. Multimodal approaches further integrate complementary morphological information across signals. Appendix Table~\ref{tab:integrated_framework} summarizes the representative modalities, modeling approaches, applications, and studies associated with each morphology.

\begin{table*}[t!]
\renewcommand{\arraystretch}{1.2}
\centering
\caption{Signal modalities and associated applications discussed in Section~\ref{sec:bio_signal_morphology}}
\label{tab:integrated_framework}
\begin{tabular}{|p{2.6cm}|p{3.2cm}|p{3.8cm}|p{5.5cm}|}
\hline
\textbf{Morphology} & \textbf{Representative Modalities} & \textbf{Modeling} & \textbf{Representative Applications \& Studies} \\ \hline

\textbf{Spikes / Bursts} &
EEG; EMG; EOG & 
Decision Tree; KNN; CNNs; Bayesian Networks; Transformers (Vision, Hybrid); Ensembles; Graph NNs; HMMs &
EEG: Brain Activity \cite{eegConformer, EEGFormer2023} \newline
EEG: Seizure Detection \cite{cai2023EEGDict, KaratasClassicalEEG2025}. \newline
EMG: Gesture Recognition \cite{lopez2024cnn, qureshi2023e2cnn, montazerin2023transformer, joby2024emg, wen2021human} \newline
EMG: Motion Prediction \cite{wang2025dual} \newline
EMG: Joint Kinematics \cite{kizyte2023influence} \newline
EMG: Blind Source Separation \cite{dere2023novel, WenEMG-CNN} \newline
EOG: Eye Movement \cite{ravichandran2021electrooculography, kim2020hidden}\newline
EOG: Sleep Staging \cite{maiti2024enhancing} \\ \hline

\textbf{Oscillatory} &
EEG; ECG; PPG &
Random Forest; Regression; SVM; SSM; CNN (Hybrid); LSTM (Bi-LSTM; Transformers  &

EEG: Sleep Staging \cite{da2017single, pouliou2025new} \newline
EEG: Emotion Recognition \cite{MahmoudEEGemotion2023}\newline
EEG: Brain State \cite{MEET2024}\newline
EEG: Schizophrenia Detection \cite{hussain2021evaluating} \newline
ECG: Arrhythmia Detection  \cite{li2022classification, kumar2023fuzz, che2021constrained, hu2022transformer, ElGhaishECGTransformer}\newline 
ECG: ECG Classification \cite{qiang2024ecgmamba}\newline 
ECG: Stress Detection \cite{AvramidisECGSSM2024}\newline 
ECG: Waveform Delineation \cite{peimankar2021dens}\newline
ECG: Blood Pressure \cite{kochev2020novel}\newline
PPG: Blood Pressure \cite{haddad2021continuous, chu2023non, ma2023kdinformer, tian2025paralleled}\newline
PPG: Heart Rate \cite{yan2025physmamba}\newline
PPG: Sleep Apnea \cite{JiangCNNApnea}\newline
PPG: Stress \cite{Hasanpoor_2022, benchekroun2022comparison} \newline
PPG: PPG Quality \cite{sivanjaneyulu2022cnn}\\ \hline

\textbf{Slow Drift} &
Pupillometry  & SVM; BiLSTM; CNNs &
Pupillometry: Attention Screening \cite{DasPupillometry, LSTM_noncog_filter}\newline
Pupillometry: PTSD \cite{taha2021detection} \\ \hline

\textbf{Hierarchical} &
ECG; EEG; Eye-tracking &
CNN+BiLSTM; Transformers; Hybrid Fourier/Transformer; Vision Transformer  &

ECG: Waveform Delineation \cite{peimankar2021dens}\newline
Eye tracking: Gaze \cite{wu2025brat, qin2025emmixformer}\newline
Eye tracking: Pupil Segmentation \cite{vayalil2024vit} \\ \hline

\textbf{Multimodal Fusion} &
EEG+EOG/EMG; ECG+PPG; EEG+Pupil/Gaze &
Hybrid CNN–Transformer; Denoising Transformers; LSTM &

Valence-Arousal: Affect Fusion \cite{Mathur_2021}\newline
Ocular: Deception Detection \cite{KhanDeception}\newline
EEG/EMG/EOG: Denoising \cite{chen2024denosieformer}\newline
EEG+pupil/eye-tracking: Depression  \cite{zhu2025transformer, zhu2025mtnet} \newline
ECG–PPG: Blood Pressure \cite{liu2024hgctnet}\newline
PPG/ECG: Reconstruction \cite{lan2023performer}\newline
ECG: Gait \cite{pham2021tfts}\newline
EEG-EMG: Motor Pattern Recognition \cite{xiao2024faconformer} \\ \hline
\end{tabular}
\end{table*}

\subsection{Ensemble and Hybrid Methodologies}
\label{sec:ensemble}
Ensemble and hybrid methodologies enhance classification performance by leveraging the complementary strengths of multiple models or feature representations, improving robustness and reducing variance. Hybrid strategies integrate handcrafted features and learned representations within a unified architecture, enabling models to capture both interpretable signal descriptors and hierarchical temporal-spatial patterns. These approaches are particularly well suited to biological signals that exhibit overlapping or mixed morphologies. EEG and ECG data frequently contain combinations of oscillatory rhythms, spatial gradients, and transient spike-like events; ensemble models that fuse temporal, spectral, and spatial features often outperform single-stream architectures in capturing this complexity~\cite{bondugula2023novel}.

Ensemble and hybrid models introduce greater computational and architectural complexity, which can complicate both training and interpretability, factors that may limit their clinical applicability. For multimodal or morphologically heterogeneous signals, these methodologies remain among the most adaptable and effective strategies for achieving high performance across diverse classification tasks.

\subsection{Datasets}
\label{sec:datasets}

To support further research into the challenges of time-series classification for biological signals, Table~\ref{tab:dataset_table} summarizes several publicly available datasets. For each dataset, we report the intended application, number of participants, associated citation, and details regarding signal modality and sampling rate. Dataset selection was guided by several criteria, including sample size, citation frequency, temporal depth, and representativeness across the modalities and morphologies discussed in this review. In recognition of the substantial interindividual variability present even within the same biosignal class, particular emphasis was placed on studies that include large participant cohorts.

Several high-volume datasets, including MIMIC-III~\cite{johnson2016mimic}, MIMIC-IV~\cite{johnson_mimic-iv_nodate}, and the UK Biobank~\cite{uk_biobank}, were originally collected with epidemiological objectives. These datasets aggregate physiological signals alongside clinical, demographic, and environmental information, enabling multimodal modeling at scale. Other large-scale resources such as Chapman ECG, Temple EEG, VitalDB, MESA, and HBN similarly provide both time-series measurements and auxiliary metadata, supporting a wide range of predictive and diagnostic tasks.

Smaller, highly cited datasets are frequently selected for their dense annotations, standardized collection protocols, or historical importance to a specific classification task. The MIT-BIH Arrhythmia Database~\cite{MoodyMITBIH}, released in 1980 with over 110,000 labeled heartbeats, has become a canonical benchmark for arrhythmia classification. Longitudinal datasets, in which participants are measured repeatedly across time, were also prioritized owing to their relevance in modeling temporal generalization and adaptation. The GRABMyo dataset~\cite{Jiang_Pradhan_He, Pradhan_He_Jiang_2022}, although smaller in scale than Ninapro~\cite{ninapro}, includes EMG recordings acquired across multiple days to capture intersession variability in gesture recognition tasks.

Datasets characterized by high-modality diversity and well-established collection procedures were also included. WESAD~\cite{Schmidt-WESAD-2018} offers multiple measurement configurations under controlled stress and amusement conditions. DEAP \cite{deap} provides synchronized recordings of multiple biosignals— including EEG, EMG, EOG, and PPG— collected under consistent experimental settings and has become a standard benchmark for affective computing.

A key limitation in comparing biological signals across datasets lies in the heterogeneity of acquisition protocols, channel configurations, and sensor placements. Even within a single modality, signal characteristics may differ substantially: ECGs can range from 1 to 12 leads, EEG systems vary in channel count and spatial layout, and EMG configurations are highly dependent on muscle group and electrode placement.

In addition to differences in signal content, the data formats, annotations, and auxiliary features vary widely across datasets. Despite the availability of large public corpora, few standardized formatting conventions have been widely adopted. One exception is the Brain Imaging Data Structure (BIDS)~\cite{bids}, originally developed for MRI and subsequently extended to EEG~\cite{eeg-bids}; BIDS is now the required standard for datasets distributed via the OpenNeuro platform~\cite{markiewicz_openneuro_2021}.

Finally, although this review focuses on biological signals, researchers may also benefit from general-purpose time-series benchmarks. The UCR archive~\cite{UCRtimeseriesarchive} includes 128 univariate datasets, while the UEA archive~\cite{UEAmultivariatetimeseries} contains 30 multivariate datasets commonly used in time-series classification research outside the biomedical domain.

\begin{table*}[]
\centering
\caption{This table provides details on several common benchmark biological signal datasets. Datasets vary significantly in number of leads/channels and sensor configuration per modality. ``––'' is listed where the sampling rate is inconsistent or unknown. For the ocular signals, E or G indicates that the data consists of EOG or gaze/pupillometry signals, respectively. Common additional signal acronyms: RR---respiratory rate measured with pneumograph or spirometry; BP---blood pressure measured with a cuff or estimated from PPG; IM---inertial measurement with accelerometer; CO$_2$---carbon dioxide concentration in exhalations measured through capnography; EDA---electrodermal activity, also known as galvanic skin response or skin conductance; and fMRI---functional magnetic resonance imaging.}
\label{tab:dataset_table}
\small
\renewcommand{\arraystretch}{1.2}
\setlength{\tabcolsep}{3pt}
\begin{tabular}{|
p{1cm}|
p{2.5cm}|
p{1.2cm}|
p{1.8cm}|
p{0.9cm}|
p{0.9cm}|
p{0.9cm}|
p{0.9cm}|
p{1.0cm}|
p{1.5cm}|}
\hline
\multicolumn{4}{|l|}{{\textbf{Dataset Details}}} &
  \multicolumn{6}{l|}{{\textbf{Signal Modality}}} \\ \hline
\multicolumn{1}{|l|}{\textbf{Name}} &
  \multicolumn{1}{l|}{\textbf{Application}} &
  \multicolumn{1}{l|}{\textbf{Participants}} &
  \textbf{Source(s)} &
  \multicolumn{1}{l|}{\textbf{EEG}} &
  \multicolumn{1}{l|}{\textbf{EMG}} &
  \multicolumn{1}{l|}{\textbf{ECG}} &
  \multicolumn{1}{l|}{\textbf{PPG}} &
  \multicolumn{1}{l|}{\textbf{Ocular}} &
  \textbf{Other} \\ \hline
\multicolumn{1}{|l|}{MIMIC} &
  \multicolumn{1}{l|}{ICU records} &
  \multicolumn{1}{l|}{\textgreater 300,000} &
  \multicolumn{1}{l|}{\cite{johnson_mimic-iii_2016}\cite{johnson_mimic-iv_nodate}} &
  \multicolumn{1}{l|}{\cellcolor[HTML]{C0C0C0}} &
  \multicolumn{1}{l|}{\cellcolor[HTML]{C0C0C0}} &
  \multicolumn{1}{l|}{––} &
  \multicolumn{1}{l|}{––} &
  \multicolumn{1}{l|}{\cellcolor[HTML]{C0C0C0}} &
  BP,RR \\ \hline
\multicolumn{1}{|l|}{UK Biobank} &
  \multicolumn{1}{l|}{Causes of disease} &
  \multicolumn{1}{l|}{\textgreater 90,000} &
  \multicolumn{1}{l|}{\cite{uk_biobank}} &
  \multicolumn{1}{l|}{\cellcolor[HTML]{C0C0C0}} &
  \multicolumn{1}{l|}{\cellcolor[HTML]{C0C0C0}} &
  \multicolumn{1}{l|}{––} &
  \multicolumn{1}{l|}{\cellcolor[HTML]{C0C0C0}} &
  \multicolumn{1}{l|}{\cellcolor[HTML]{C0C0C0}} &
  IM \\ \hline
\multicolumn{1}{|l|}{Chapman ECG} &
  \multicolumn{1}{l|}{Arrhythmia detection} &
  \multicolumn{1}{l|}{45,152} &
  \multicolumn{1}{l|}{\cite{zheng_large_nodate}\cite{zheng_optimal_2020}} &
  \multicolumn{1}{l|}{\cellcolor[HTML]{C0C0C0}} & 
  \multicolumn{1}{l|}{\cellcolor[HTML]{C0C0C0}} & 
  \multicolumn{1}{l|}{500 Hz} & 
  \multicolumn{1}{l|}{\cellcolor[HTML]{C0C0C0}} & 
  \multicolumn{1}{l|}{\cellcolor[HTML]{C0C0C0}} & 
  \multicolumn{1}{l|}{\cellcolor[HTML]{C0C0C0}} \\ \hline 
\multicolumn{1}{|l|}{PTB-XL} &
  \multicolumn{1}{l|}{ECG interpretation} &
  \multicolumn{1}{l|}{18,869} &
  \multicolumn{1}{l|}{\cite{wagner_ptb-xl_2020}\cite{wagner_ptb-xl_nodate}} &
  \multicolumn{1}{l|}{\cellcolor[HTML]{C0C0C0}} &
  \multicolumn{1}{l|}{\cellcolor[HTML]{C0C0C0}} &
  \multicolumn{1}{l|}{500 Hz} &
  \multicolumn{1}{l|}{\cellcolor[HTML]{C0C0C0}} &
  \multicolumn{1}{l|}{\cellcolor[HTML]{C0C0C0}} &
  \multicolumn{1}{l|}{\cellcolor[HTML]{C0C0C0}} \\ \hline
\multicolumn{1}{|l|}{Temple EEG Corpus} &
  \multicolumn{1}{l|}{EEG diagnostics} & 
  \multicolumn{1}{l|}{10,874} & 
  \multicolumn{1}{l|}{\cite{obeid2016temple}} & 
  \multicolumn{1}{l|}{250 Hz} & 
  \multicolumn{1}{l|}{\cellcolor[HTML]{C0C0C0}} & 
  \multicolumn{1}{l|}{\cellcolor[HTML]{C0C0C0}} & 
  \multicolumn{1}{l|}{\cellcolor[HTML]{C0C0C0}} & 
  \multicolumn{1}{l|}{\cellcolor[HTML]{C0C0C0}} & 
  \multicolumn{1}{l|}{\cellcolor[HTML]{C0C0C0}} \\ \hline 
\multicolumn{1}{|l|}{Physionet/CINC 2017} &
  \multicolumn{1}{l|}{1-lead AFib detection} &
  \multicolumn{1}{l|}{8,528} &
  \multicolumn{1}{l|}{\cite{clifford_af_2017}} &
  \multicolumn{1}{l|}{\cellcolor[HTML]{C0C0C0}} &
  \multicolumn{1}{l|}{\cellcolor[HTML]{C0C0C0}} &
  \multicolumn{1}{l|}{300 Hz} &
  \multicolumn{1}{l|}{\cellcolor[HTML]{C0C0C0}} &
  \multicolumn{1}{l|}{\cellcolor[HTML]{C0C0C0}} &
  \multicolumn{1}{l|}{\cellcolor[HTML]{C0C0C0}} \\ \hline
\multicolumn{1}{|l|}{VitalDB} &
  \multicolumn{1}{l|}{Surgical vitals} &
  \multicolumn{1}{l|}{6,388} &
  \multicolumn{1}{l|}{\cite{lee_vitaldb_2022}} &
  \multicolumn{1}{l|}{128 Hz} &
  \multicolumn{1}{l|}{\cellcolor[HTML]{C0C0C0}} &
  \multicolumn{1}{l|}{500 Hz} &
  \multicolumn{1}{l|}{500 Hz} &
  \multicolumn{1}{l|}{\cellcolor[HTML]{C0C0C0}} &
  \multicolumn{1}{l|}{BP,CO$_2$} \\ \hline
\multicolumn{1}{|l|}{MESA} &
  \multicolumn{1}{l|}{Atherosclerosis study} &
  \multicolumn{1}{l|}{6,765} &
  \multicolumn{1}{l|}{\cite{mesa_2002}\cite{mesa_chen_2015}} &
  \multicolumn{1}{l|}{––} & 
  \multicolumn{1}{l|}{––} & 
  \multicolumn{1}{l|}{––} & 
  \multicolumn{1}{l|}{––} & 
  \multicolumn{1}{l|}{E ––} & 
  \multicolumn{1}{l|}{BP,RR} \\ \hline 
\multicolumn{1}{|l|}{HBN} &
  \multicolumn{1}{l|}{Pediatric mental health} &
  \multicolumn{1}{l|}{\textgreater 3,000} &
  \multicolumn{1}{l|}{\cite{alexander_open_2017}} &
  \multicolumn{1}{l|}{500 Hz} &
  \multicolumn{1}{l|}{\cellcolor[HTML]{C0C0C0}} &
  \multicolumn{1}{l|}{\cellcolor[HTML]{C0C0C0}} &
  \multicolumn{1}{l|}{\cellcolor[HTML]{C0C0C0}} &
  \multicolumn{1}{l|}{G 120 Hz} &
  fMRI \\ \hline
\multicolumn{1}{|l|}{GazeBaseVR} &
  \multicolumn{1}{l|}{Eye tracking} &
  \multicolumn{1}{l|}{407} &
  \multicolumn{1}{l|}{\cite{lohr_gazebasevr_2023}} &
  \multicolumn{1}{l|}{\cellcolor[HTML]{C0C0C0}} & 
  \multicolumn{1}{l|}{\cellcolor[HTML]{C0C0C0}} & 
  \multicolumn{1}{l|}{\cellcolor[HTML]{C0C0C0}} & 
  \multicolumn{1}{l|}{\cellcolor[HTML]{C0C0C0}} & 
  \multicolumn{1}{l|}{G 250 Hz} & 
  \multicolumn{1}{l|}{\cellcolor[HTML]{C0C0C0}} \\ \hline 
\multicolumn{1}{|l|}{GazeBase} &
  \multicolumn{1}{l|}{Eye tracking} &
  \multicolumn{1}{l|}{322} &
  \multicolumn{1}{l|}{\cite{gazebase_2021}} &
  \multicolumn{1}{l|}{\cellcolor[HTML]{C0C0C0}} & 
  \multicolumn{1}{l|}{\cellcolor[HTML]{C0C0C0}} & 
  \multicolumn{1}{l|}{\cellcolor[HTML]{C0C0C0}} & 
  \multicolumn{1}{l|}{\cellcolor[HTML]{C0C0C0}} & 
  \multicolumn{1}{l|}{G 1,000 Hz} & 
  \multicolumn{1}{l|}{\cellcolor[HTML]{C0C0C0}} \\ \hline 
\multicolumn{1}{|l|}{Ninapro} &
  \multicolumn{1}{l|}{Neuroprosthetics} &
  \multicolumn{1}{l|}{264} &
  \multicolumn{1}{l|}{\cite{ninapro}} &
  \multicolumn{1}{l|}{\cellcolor[HTML]{C0C0C0}} & 
  \multicolumn{1}{l|}{100 Hz} & 
  \multicolumn{1}{l|}{\cellcolor[HTML]{C0C0C0}} & 
  \multicolumn{1}{l|}{\cellcolor[HTML]{C0C0C0}} & 
  \multicolumn{1}{l|}{\cellcolor[HTML]{C0C0C0}} & 
  \multicolumn{1}{l|}{\cellcolor[HTML]{C0C0C0}} \\ \hline 
\multicolumn{1}{|l|}{Montreal Archive} &
  \multicolumn{1}{l|}{Sleep study} &
  \multicolumn{1}{l|}{200} &
  \multicolumn{1}{l|}{\cite{montreal_sleep}} &
  \multicolumn{1}{l|}{256 Hz} & 
  \multicolumn{1}{l|}{256 Hz} & 
  \multicolumn{1}{l|}{––} & 
  \multicolumn{1}{l|}{––} & 
  \multicolumn{1}{l|}{E 256 Hz} & 
  \multicolumn{1}{l|}{EDA,RR} \\ \hline 
\multicolumn{1}{|l|}{Sleep-EDF} &
  \multicolumn{1}{l|}{Sleep study} &
  \multicolumn{1}{l|}{197} &
  \multicolumn{1}{l|}{\cite{SleepEEG}} &
  \multicolumn{1}{l|}{100 Hz} & 
  \multicolumn{1}{l|}{100 Hz} & 
  \multicolumn{1}{l|}{\cellcolor[HTML]{C0C0C0}} & 
  \multicolumn{1}{l|}{\cellcolor[HTML]{C0C0C0}} & 
  \multicolumn{1}{l|}{E 100 Hz} & 
  \multicolumn{1}{l|}{\cellcolor[HTML]{C0C0C0}} \\ \hline 
\multicolumn{1}{|l|}{EEG MMI} &
  \multicolumn{1}{l|}{Motor imagery} &
  \multicolumn{1}{l|}{109} &
  \multicolumn{1}{l|}{\cite{eeg_mmi}} &
  \multicolumn{1}{l|}{160 Hz} & 
  \multicolumn{1}{l|}{\cellcolor[HTML]{C0C0C0}} & 
  \multicolumn{1}{l|}{\cellcolor[HTML]{C0C0C0}} & 
  \multicolumn{1}{l|}{\cellcolor[HTML]{C0C0C0}} & 
  \multicolumn{1}{l|}{\cellcolor[HTML]{C0C0C0}} & 
  \multicolumn{1}{l|}{\cellcolor[HTML]{C0C0C0}} \\ \hline 
\multicolumn{1}{|l|}{EEGdenoiseNet} &
  \multicolumn{1}{l|}{Denoising} &
  \multicolumn{1}{l|}{105} &
  \multicolumn{1}{l|}{\cite{ninapro}} &
  \multicolumn{1}{l|}{256 Hz} & 
  \multicolumn{1}{l|}{512 Hz} & 
  \multicolumn{1}{l|}{\cellcolor[HTML]{C0C0C0}} & 
  \multicolumn{1}{l|}{\cellcolor[HTML]{C0C0C0}} & 
  \multicolumn{1}{l|}{E 256 Hz} & 
  \multicolumn{1}{l|}{\cellcolor[HTML]{C0C0C0}} \\ \hline 
\multicolumn{1}{|l|}{MODMA} &
  \multicolumn{1}{l|}{Mental disorder analys.} &
  \multicolumn{1}{l|}{84} &
  \multicolumn{1}{l|}{\cite{modma_dataset}} &
  \multicolumn{1}{l|}{250 Hz} & 
  \multicolumn{1}{l|}{\cellcolor[HTML]{C0C0C0}} & 
  \multicolumn{1}{l|}{\cellcolor[HTML]{C0C0C0}} & 
  \multicolumn{1}{l|}{\cellcolor[HTML]{C0C0C0}} & 
  \multicolumn{1}{l|}{\cellcolor[HTML]{C0C0C0}} & 
  \multicolumn{1}{l|}{Audio} \\ \hline 
\multicolumn{1}{|l|}{Wainstein ADHD} &
  \multicolumn{1}{l|}{ADHD diagnosis} &
  \multicolumn{1}{l|}{50} &
  \multicolumn{1}{l|}{\cite{wainstein_data}} &
  \multicolumn{1}{l|}{\cellcolor[HTML]{C0C0C0}} & 
  \multicolumn{1}{l|}{\cellcolor[HTML]{C0C0C0}} & 
  \multicolumn{1}{l|}{\cellcolor[HTML]{C0C0C0}} & 
  \multicolumn{1}{l|}{\cellcolor[HTML]{C0C0C0}} & 
  \multicolumn{1}{l|}{G 1,000 Hz} & 
  \multicolumn{1}{l|}{\cellcolor[HTML]{C0C0C0}} \\ \hline 
\multicolumn{1}{|l|}{MIT-BIH} &
  \multicolumn{1}{l|}{ECG diagnostic} &
  \multicolumn{1}{l|}{47} &
  \multicolumn{1}{l|}{\cite{MoodyMITBIH}} &
  \multicolumn{1}{l|}{\cellcolor[HTML]{C0C0C0}} & 
  \multicolumn{1}{l|}{\cellcolor[HTML]{C0C0C0}} & 
  \multicolumn{1}{l|}{360 Hz} & 
  \multicolumn{1}{l|}{\cellcolor[HTML]{C0C0C0}} & 
  \multicolumn{1}{l|}{\cellcolor[HTML]{C0C0C0}} & 
  \multicolumn{1}{l|}{\cellcolor[HTML]{C0C0C0}} \\ \hline 
\multicolumn{1}{|l|}{Vortal} &
  \multicolumn{1}{l|}{RR estimation} &
  \multicolumn{1}{l|}{45} &
  \multicolumn{1}{l|}{\cite{charlton_assessment_2016}} &
  \multicolumn{1}{l|}{\cellcolor[HTML]{C0C0C0}} & 
  \multicolumn{1}{l|}{\cellcolor[HTML]{C0C0C0}} & 
  \multicolumn{1}{l|}{500 Hz} & 
  \multicolumn{1}{l|}{500 Hz} & 
  \multicolumn{1}{l|}{\cellcolor[HTML]{C0C0C0}} & 
  \multicolumn{1}{l|}{RR} \\ \hline 
\multicolumn{1}{|l|}{GRABMyo} &
  \multicolumn{1}{l|}{Gesture recognition} &
  \multicolumn{1}{l|}{43} &
  \multicolumn{1}{l|}{\cite{Jiang_Pradhan_He}\cite{Pradhan_He_Jiang_2022}} &
  \multicolumn{1}{l|}{\cellcolor[HTML]{C0C0C0}} & 
  \multicolumn{1}{l|}{2,048 Hz} & 
  \multicolumn{1}{l|}{\cellcolor[HTML]{C0C0C0}} & 
  \multicolumn{1}{l|}{\cellcolor[HTML]{C0C0C0}} & 
  \multicolumn{1}{l|}{\cellcolor[HTML]{C0C0C0}} & 
  \multicolumn{1}{l|}{\cellcolor[HTML]{C0C0C0}} \\ \hline 
\multicolumn{1}{|l|}{Capnobase} &
  \multicolumn{1}{l|}{RR benchmark} &
  \multicolumn{1}{l|}{42} &
  \multicolumn{1}{l|}{\cite{karlen_capnobase_2021}\cite{capnobase_paper}} &
  \multicolumn{1}{l|}{\cellcolor[HTML]{C0C0C0}} & 
  \multicolumn{1}{l|}{\cellcolor[HTML]{C0C0C0}} & 
  \multicolumn{1}{l|}{\cellcolor[HTML]{C0C0C0}} & 
  \multicolumn{1}{l|}{100 Hz} & 
  \multicolumn{1}{l|}{\cellcolor[HTML]{C0C0C0}} & 
  \multicolumn{1}{l|}{CO$_2$} \\ \hline 
\multicolumn{1}{|l|}{DEAP} &
  \multicolumn{1}{l|}{Emotion recognition} &
  \multicolumn{1}{l|}{32} &
  \multicolumn{1}{l|}{\cite{deap}} &
  \multicolumn{1}{l|}{256 Hz} & 
  \multicolumn{1}{l|}{256 Hz} & 
  \multicolumn{1}{l|}{256 Hz} & 
  \multicolumn{1}{l|}{256 Hz} & 
  \multicolumn{1}{l|}{E 256 Hz} & 
  \multicolumn{1}{l|}{EDA,RR} \\ \hline 
\multicolumn{1}{|l|}{SEED-VII} &
  \multicolumn{1}{l|}{Emotion recognition} &
  \multicolumn{1}{l|}{20} &
  \multicolumn{1}{l|}{\cite{seed-VII_dataset}} &
  \multicolumn{1}{l|}{200 Hz} & 
  \multicolumn{1}{l|}{\cellcolor[HTML]{C0C0C0}} & 
  \multicolumn{1}{l|}{\cellcolor[HTML]{C0C0C0}} & 
  \multicolumn{1}{l|}{\cellcolor[HTML]{C0C0C0}} & 
  \multicolumn{1}{l|}{\cellcolor[HTML]{C0C0C0}} & 
  \multicolumn{1}{l|}{\cellcolor[HTML]{C0C0C0}} \\ \hline 
\multicolumn{1}{|l|}{Hyser} &
  \multicolumn{1}{l|}{Multitask benchmark} &
  \multicolumn{1}{l|}{20} &
  \multicolumn{1}{l|}{\cite{hyser_dataset}\cite{hyser_paper}} &
  \multicolumn{1}{l|}{\cellcolor[HTML]{C0C0C0}} & 
  \multicolumn{1}{l|}{2,048 Hz} & 
  \multicolumn{1}{l|}{\cellcolor[HTML]{C0C0C0}} & 
  \multicolumn{1}{l|}{\cellcolor[HTML]{C0C0C0}} & 
  \multicolumn{1}{l|}{\cellcolor[HTML]{C0C0C0}} & 
  \multicolumn{1}{l|}{\cellcolor[HTML]{C0C0C0}} \\ \hline 
\multicolumn{1}{|l|}{BCI IV} &
  \multicolumn{1}{l|}{Motor imagery} &
  \multicolumn{1}{l|}{16} &
  \multicolumn{1}{l|}{\cite{brunner2008bci}\cite{blankertz2007non}} &
  \multicolumn{1}{l|}{1,000 Hz} & 
  \multicolumn{1}{l|}{\cellcolor[HTML]{C0C0C0}} & 
  \multicolumn{1}{l|}{\cellcolor[HTML]{C0C0C0}} & 
  \multicolumn{1}{l|}{\cellcolor[HTML]{C0C0C0}} & 
  \multicolumn{1}{l|}{E 250 Hz} & 
  \multicolumn{1}{l|}{\cellcolor[HTML]{C0C0C0}} \\ \hline 
\multicolumn{1}{|l|}{WESAD} &
  \multicolumn{1}{l|}{Stress/affect detection} &
  \multicolumn{1}{l|}{15} &
  \multicolumn{1}{l|}{\cite{Schmidt-WESAD-2018}} &
  \multicolumn{1}{l|}{\cellcolor[HTML]{C0C0C0}} & 
  \multicolumn{1}{l|}{700 Hz} & 
  \multicolumn{1}{l|}{700 Hz} & 
  \multicolumn{1}{l|}{700 Hz} & 
  \multicolumn{1}{l|}{\cellcolor[HTML]{C0C0C0}} & 
  \multicolumn{1}{l|}{EDA,IM} \\ \hline 
\end{tabular}
\end{table*}

\subsection{Miscellaneous Signals}
\label{sec:misc}
Several additional biological signals fall outside the primary scope of this review but remain relevant to time-series classification research. These include gait, respiration, facial expressions, and physical gestures. Such modalities often exhibit complex, nonoscillatory temporal structures and are frequently captured through external, behaviorally linked measurement systems, in contrast to the primarily internal physiological signals emphasized in this work.

Gait dynamics exhibit hierarchical rhythms are widely used for activity recognition and fall detection~\cite{cicirelli2022human}. Respiratory signals display slow drift and low-frequency oscillations, supporting applications in sleep staging and stress monitoring~\cite{vanegas2020sensing}. Facial expression dynamics captured via surface electromyography employed in emotion recognition contain bursts of muscular activity~\cite{tian2011facial}, whereas gesture signals recorded with wearable sensors exhibit hierarchical rhythms enable real-time human–computer interaction~\cite{vasconez2022hand}. In the context of deception detection, temporal patterns capturing spikes in facial microexpressions have incorporated SVM probabilities as input features improved classification performance from 65\% to between 82\% and 90\% accuracy~\cite{WuDeceptionVideo}, highlighting the utility of combining interpretable temporal features with discriminative learning models in behaviorally expressive signal domains.

\bibliographystyle{IEEEtran}
\bibliography{ref}

\end{document}